\documentclass[acmtog,nonacm]{acmart}

\usepackage{booktabs} 
\usepackage{bm}
\usepackage{multirow}
\usepackage{multicol}
\usepackage{ulem}
\usepackage{color}
\definecolor{Red}{cmyk}{0,1,1,0}
\definecolor{Green}{cmyk}{1,0,1,0}
\definecolor{Cyan}{cmyk}{1,0,0,0}
\definecolor{Purple}{cmyk}{0.45,0.86,0,0}
\definecolor{Sky}{cmyk}{1.0,0.8,0,0}
\definecolor{Rosolic}{cmyk}{0.00,1.00,0.50,0}
\definecolor{Blue}{cmyk}{1.00,1.00,0.00,0}
\definecolor{Orange}{cmyk}{0,0.52,0.80,0}
\definecolor{Black}{cmyk}{1,0,0,1}
\newcommand{\dzh}{\textcolor{Black}}

\usepackage{xcolor}

\usepackage[ruled]{algorithm2e} 

\SetAlFnt{\small}
\SetAlCapFnt{\small}
\SetAlCapNameFnt{\small}
\SetAlCapHSkip{0pt}

\acmJournal{TOG}

\begin{document}
\title{Towards Better Robustness: Pose-Free 3D Gaussian Splatting for Arbitrarily Long Videos}

\author{Zhen-Hui Dong}
\authornotemark[1]
\email{dzh23@mails.tsinghua.edu.cn}
\author{Sheng Ye}
\authornote{These authors contributed equally to this work.}
\email{yec22@mails.tsinghua.edu.cn}
\affiliation{
 \institution{Tsinghua University}
 \city{Beijing}
 \country{China}
}

\author{Yu-Hui Wen}
\affiliation{
 \institution{Beijing Jiaotong University}
 \city{Beijing}
 \country{China}
}
\email{yhwen1@bjtu.edu.cn}

\author{Nannan Li}
\affiliation{
 \institution{Maritime University}
 \city{DaLian}
 \country{China}
}
\email{nannanli@dlmu.edu.cn}

\author{Yong-Jin Liu}
\authornote{Corresponding authors.}
\affiliation{
 \institution{Tsinghua University}
 \city{Beijing}
 \country{China}
}
\email{liuyongjin@tsinghua.edu.cn}

\begin{abstract}
3D Gaussian Splatting (3DGS) has emerged as a powerful representation due to its efficiency and high-fidelity rendering. 3DGS training requires a known camera pose for each input view, typically obtained by Structure-from-Motion (SfM) pipelines. Pioneering works have attempted to relax this restriction but still face difficulties when handling long sequences with complex camera trajectories. In this paper, we propose Rob-GS, a robust framework to progressively estimate camera poses and optimize 3DGS for arbitrarily long video inputs. In particular, by leveraging the inherent continuity of videos, we design an adjacent pose tracking method to ensure stable pose estimation between consecutive frames. To handle arbitrarily long inputs, we propose a Gaussian visibility retention check strategy to adaptively split the video sequence into several segments and optimize them separately. Extensive experiments on \textit{Tanks and Temples}, \textit{ScanNet}, and a self-captured dataset show that Rob-GS outperforms the state-of-the-arts.
\end{abstract}

%
%

%
%


\begin{teaserfigure}
  \setlength{\abovecaptionskip}{4pt}  
  \setlength{\belowcaptionskip}{0pt}
  \centering
  \includegraphics[width=\textwidth]{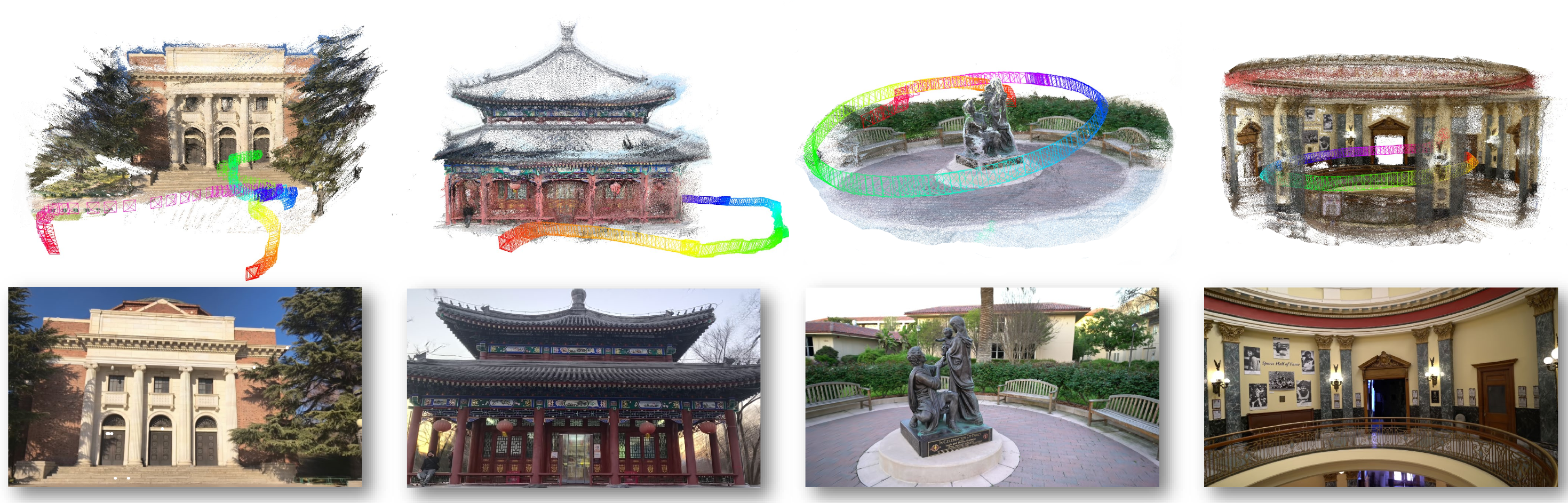}
  \caption{
  \textbf{Camera pose estimations and novel view synthesis}. We propose Rob-GS to reconstruct scenes from arbitrarily long, casually captured videos without known camera poses. Rob-GS enables robust pose estimations (first row) and high-quality renderings (second row).
  }
  \label{fig:teaser}
  \Description{teaser}
  \vspace{0.2cm}
\end{teaserfigure}

\maketitle

\section{Introduction}
Photo-realistic novel view synthesis is a fundamental problem in computer vision and graphics, and has been significantly advanced by the emergence of Neural Radiance Field~ (NeRF)~\cite{nerf} and 3D Gaussian Splatting~(3DGS)~\cite{3DGS}.
An essential pre-processing step for existing NeRF and 3DGS techniques is to obtain the camera poses for all input images, which is usually achieved by Structure-from-Motion (SfM) methods such as COLMAP~\cite{colmap}.
However, SfM becomes time-consuming with an increasing number of images, and struggles in textureless regions. Moreover, SfM frequently fails under casually captured video inputs due to varying lighting, unstable motion, complex scene layouts, and random camera trajectories, limiting its applicability in real-world scenarios.

To eliminate the need for pre-computing camera poses by SfM methods, some pioneering works~\cite{barf,SC-nerf,nerfmm,nope-nerf} have been proposed to optimize camera poses and NeRFs jointly. Nevertheless, these NeRF-based methods are limited by the computational inefficiency during training and rendering.
Subsequently, CF-3DGS~\cite{CF3dgs} first proposes to train SfM-free 3DGS, and shows superior performance in rendering quality, pose estimation accuracy, and training speed.
However, it is only effective for short input sequences where adjacent views do not change significantly.
When dealing with more challenging inputs (\textit{e.g.,} long sequences 
and low-overlap frames), CF-3DGS 
produces inaccurate poses and degraded renderings.

High-fidelity reconstruction of large-scale scenes necessitates long input videos (containing hundreds to thousands of frames) with complex camera trajectories (covering various regions). 
In this scenario, there remain two core challenges:
(1) robust pose estimation for long and complex camera trajectories and (2) memory-efficient 3DGS-based reconstruction of large scenes without overflow.

To address the above challenges, we propose \textbf{Rob}ust Pose-Free 3D \textbf{G}aussian \textbf{S}platting (Rob-GS) in this work.
Rob-GS introduces the following two key components:
(1) An adjacent pose tracking method that leverages the inherent temporal continuity of videos. 
The adjacent camera pose transformations between consecutive frames are relatively small and easy to learn.
Specifically, we first fit current frame using a set of Gaussians, then freeze them and optimize the camera pose transformation until these Gaussians can precisely render the subsequent frame. We incorporate the optimization into 3DGS CUDA-rasterizer for better efficiency.
To further enhance the pose estimation robustness, we introduce another constraint, forcing the projection flow derived from depths and camera poses to match with the optical flow;
(2) An adaptive segmentation strategy to handle arbitrarily long inputs. 
We split the video sequence into segments and optimize them separately. Considering that distant frames may no longer represent the same region due to the long camera trajectory, 
our strategy checks Gaussian visibility retention, ensuring each segment focuses on a coherent portion of the scene.
These two components guarantee stable pose estimation and high-quality rendering while preventing memory overflow under challenging situations.
Compared to the existing baselines, our method performs robustly on much more complex and longer input sequences, taking a step towards practical deployment.
 
We evaluate the performance of Rob-GS on widely used benchmark datasets~\cite{tanks, scannet}.
In contrast to previous works~\cite{barf,nope-nerf,CF3dgs} that only validate on a small subset of input video sequences (restricted to small regions), we utilize much longer sequences (covering large areas) for evaluation.
Furthermore, we also evaluate the robustness of our method on a self-captured dataset under complicated real-world conditions, where videos are casually captured (single-pass, unconstrained manner) using a mobile phone without any stabilizer.
In summary, we make the following contributions:
\begin{itemize}
\item We introduce a novel framework, Rob-GS, to progressively estimate camera poses and optimize 3DGS for arbitrarily long input video sequences.
\item We design a robust flow-induced adjacent pose estimation method to effectively handle challenging, casually captured camera trajectories (\textit{e.g.,} with large or fast camera motion).
\item We propose an adaptive segmentation strategy based on Gaussian visibility retention to process long inputs, maintaining rendering quality while avoiding memory overflow.
\item Extensive experiments show that Rob-GS outperforms state-of-the-art methods in terms of rendering quality, pose estimation accuracy, and training speed.
\end{itemize}

\begin{figure*}
    \centering
    \setlength{\abovecaptionskip}{4pt}  
    \setlength{\belowcaptionskip}{0pt}
    \includegraphics[width=0.95\linewidth]{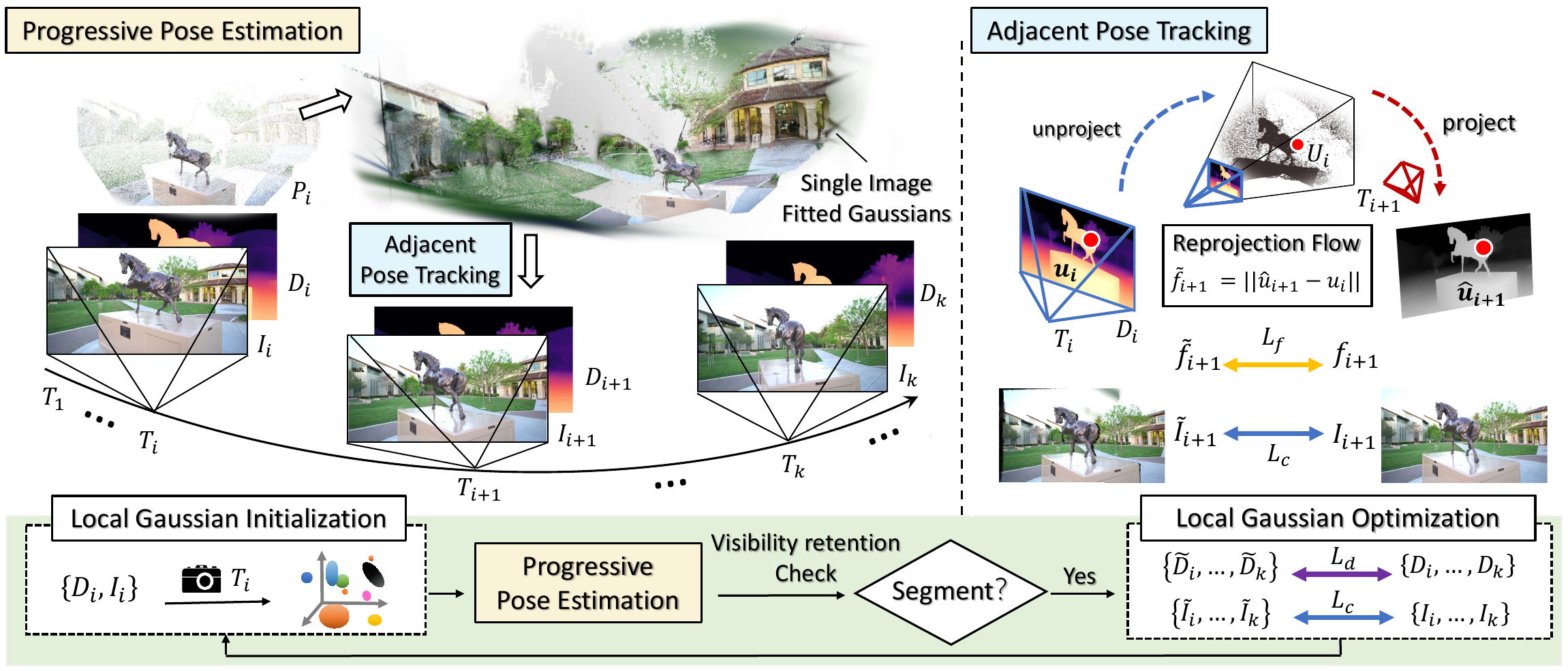}
    \caption{\textbf{The overall framework of Rob-GS}. We use 3D Gaussians (Sec.~\ref{sec:3dgs}) as the scene representation, and progressively estimate camera poses using a robust tracking approach (Sec.~\ref{sec:pose}) that leverages adjacent image pairs. To handle long video sequence, we design an adaptive segmentation scheme (Sec.~\ref{sec:GS optimization}) to split the video sequence into several local segments and optimize them individually.}
    \Description{pipeline}
    \label{fig:pipeline}
\end{figure*}

\section{Related Works}
\subsection{Novel View Synthesis}
Novel view synthesis aims to generate images from unseen camera views.
Traditional methods are based on light fields~\cite{2001unstructuredlumigraph,local-lightfield} or multi-plane images (MPIs)~\cite{stereo-mpi,single-mpi}.
NeRF~\cite{nerf} represents 3D scenes as implicit fields  
encoded within an MLP network, and synthesizes novel views through volume rendering. Although various works are proposed to improve NeRF's performance~\cite{nerf++,mip-nerf,Ref-nerf,instant-ngp,plenoxel,tensorf,block-nerf,zip-nerf}, NeRF and its variants are still limited by the inability to render in real time.
To address this challenge, 3DGS~\cite{3DGS} employs a differentiable point-based splatting method that achieves high-fidelity and real-time rendering. Subsequent works build upon 3DGS to deal with anti-aliasing~\cite{mip-splatting}, compression~\cite{compact-gs}, inconsistency~\cite{gaussian-dk}, city-scale scenes~\cite{citygaussian}, dynamic scenes~\cite{4d-gs}, etc.
However, most of these studies still rely on pre-computed camera parameters.
\subsection{Pose-Free Novel View Synthesis}
Both NeRF-based and 3DGS-based approaches have been attempted to eliminate the need for known or provided camera poses.
Specifically, i-NeRF~\cite{inerf} proposes to estimate camera poses by matching keypoints using a pre-trained NeRF.
Methods like NeRFmm~\cite{nerfmm}, BARF~\cite{barf}, and SC-NeRF~\cite{SC-nerf} simultaneously optimize poses and NeRF. Yet, they lack robustness as they are limited to forward-facing scenes or heavily rely on good initialization poses.
To handle complicated scenes with larger camera motion, Nope-NeRF~\cite{nope-nerf} uses monocular estimated depths to constrain the poses and geometry of NeRF.
ICON~\cite{icon} introduces a confidence-based optimization scheme and removes the requirement for pose initialization, but mainly focuses on object-centric scenarios.
LocalRF~\cite{localrf} solves long and complex trajectories by dynamically allocating new local radiance fields, significantly improving robustness but requiring a long training time.

For 3DGS without SfM, CF-3DGS~\cite{CF3dgs} learns rigid transformations of 3D Gaussians to approximate the relative camera movements, outperforming previous NeRF-based methods. However, it performs well only on short image sequences and frequently fails in casually captured videos.
This limitation stems from its pose estimation mechanism, which lacks robustness under large inter-frame motion.
Moreover, under long input sequences, the incremental optimization strategy of CF-3DGS cannot maintain global consistency, and suffers from excessive Gaussian densification, resulting in out-of-memory issue.
Concurrently, Free-SurGS~\cite{Free-SurGS} focuses on tackling pose-free surgical scene reconstruction.
Building on 3DGS, we aim to develop a robust framework for pose-free novel view synthesis, specifically designed for challenging scenarios: casually captured videos characterized by complex trajectories, large camera motion, and arbitrarily long sequences.

\section{Method}

\subsection{Overall Pipeline}
We propose Rob-GS, a robust framework (see Fig.~\ref{fig:pipeline}) for novel view synthesis capable of handling arbitrarily long input video sequences, without the need for tedious SfM pre-processing.
Our framework progressively estimates camera poses and optimizes the 3D scene.
Specifically, we propose an adaptive sementation strategy that splits the video sequence into several overlapping segments and optimizes them separately. These designs ensure the stability of optimization and prevent memory overflow under long video inputs.

Starting from the first video frame, we progressively estimate camera poses using a robust tracking approach that leverages each adjacent image pair. 
After tracking a new frame, we conduct a simple yet effective visibility retention check to determine whether the tracked frames should initiate a new segment.
If a new segment is warranted, we perform local optimization; otherwise (or after local optimization), pose estimation proceeds.
By estimating camera poses globally across the entire video, we place all local segments within a unified coordinate system, maintaining geometric consistency throughout the scene.
In the following sections, we first briefly review the 3D Gaussian representation (Sec.~\ref{sec:3dgs}), then introduce our robust pose estimation method (Sec.~\ref{sec:pose}), and finally describe the segmentation strategy and local 3DGS optimization (Sec.~\ref{sec:GS optimization}).

\subsection{Preliminary on 3D Gaussian Representation}
\label{sec:3dgs}
Rob-GS presents the scene using numerous anisotropic 3D Gaussians~\cite{3DGS}. Each Gaussian is associated with geometric and appearance properties, which include position $\mu$, covariance $\Sigma$, opacity $\alpha$, and color $c$. The covariance can be further decomposed into a scaling matrix and a rotation matrix: \(\Sigma=RSS^TR^T\). The 3D Gaussian at
any spatial point $x$ can then be defined as:
\begin{equation}
G(x) = e^{-\frac{1}{2} (x - \mu)^T \Sigma^{-1} (x - \mu)}.   
\end{equation}
Given a camera pose \(T_{WC} \in SE(3)\), 3D Gaussians are first projected to 2D Gaussians for rendering. The mean $\mu'$ and covariance $\Sigma'$ of each 2D Gaussian is computed as follows: 
\begin{equation}
    \mu' = \pi(T_{WC} \cdot \mu), \ 
    \Sigma' = J W \Sigma W^T J^T,
\label{eq:2}
\end{equation}
where $\pi$ is the projection operation, $J$ is the Jacobian of the affine approximation of the projective transformation, and $W$ is the rotational component of $T_{WC}$. Following the point-based differential rendering, the color and depth of one pixel are obtained by $\alpha$-blending:
\begin{equation}
\resizebox{0.7\linewidth}{!}{$
\displaystyle
    C = \sum_{i \in N} c_i \alpha_i' \prod_{j=1}^{i-1} (1 - \alpha_j'), \ 
    D = \sum_{i \in N} d_i \alpha_i' \prod_{j=1}^{i-1} (1 - \alpha_j').
$}
\end{equation}
where $c_i$ represents the color of the $i$-th 3D Gaussian, $d_i$ is the depth of the $i$-th Gaussian in camera coordinates, and $\alpha_i'$ is the opacity weighted by 2D Gaussian with the covariance $\Sigma'$.

In Kerbl \textit{et al.}~\shortcite{3DGS}, 3D Gaussians are initialized from a sparse SfM~\cite{colmap} point cloud with calibrated camera poses.
In contrast, our method progressively recovers the camera poses and optimizes Gaussian properties jointly.

\subsection{Robust Pose Estimation}
\label{sec:pose}
To achieve stable and robust pose estimation, we utilize the inherent properties of video streams, namely, the sequential frame ordering and significant overlap between adjacent frames.
To this end, we fit a set of 3D Gaussians $G_i$ to each frame $I_i$, and use $G_i$ to estimate the pose of subsequent frame $I_{i+1}$. For the initial frame $I_0$, we directly set its camera pose to an identity matrix.
Given a new frame $I_{i+1}$, its corresponding camera pose $T_{i+1}$ is initialized as the previous pose $T_{i}$.
We then optimize $T_{i+1}$ by rendering the image $\tilde{I}_{i+1}$ under $T_{i+1}$ using $G_{i}$, and minimizing the color difference between $\tilde{I}_{i+1}$ and the ground-truth frame $I_{i+1}$. In addition, we introduce a flow loss (Sec.~\ref{sec:pose-flow}) to complement the color supervision, aiming to improve robustness under large viewpoint changes between frames.

\subsubsection{Single Image Fitted Gaussians}
As illustrated in Fig.~\ref{fig:pipeline}, at timestep $i$, we first estimate the monocular depth $D_i$ by an off-the-shelf model~\cite{depth_anything}. Then, we unproject frame $I_i$ using its estimated depth $D_i$ to generate a colored point cloud $P_i$.
We initialize Gaussians $G_i$ from $P_i$ and optimize all Gaussian attributes $\theta$ to minimize the color and depth difference:
\begin{equation}
    G_{i}(\theta^{*}) = \arg \min_{\theta} \left\{ L_{c} (\tilde{I_i}, I_{i}) + \lambda_d L_{d} \right\},
\end{equation}
where $\tilde{I_i}$ is the rendered image, $L_{c}$ is the $L_{1}$ loss combined with SSIM loss, $L_{d}$ is the scale-invariant depth loss:
\begin{equation}
    L_{d} = | | (w \tilde{D_i} + q) -  D_{i} | | ^{2},
\end{equation}
Here, $w$ and $q$ are the scale and shift factors used to align rendered depth $\tilde{D_i}$ and monocular depth $D_i$, computed by a least-squares criterion~\cite{eigen2014depthloss,ranftl2020depthloss}.
Note that in this fitting process, we do not conduct any Gaussian pruning or densification.

\subsubsection{Flow-Induced Adjacent Pose Tracking}
\label{sec:pose-flow}
We design a flow loss to guide the pose estimation using dense correspondences computed via reprojection flow and optical flow.
The reprojection flow calculates the per-pixel movement between adjacent frames.
Given a 2D pixel $u_i \in \mathbb{R}^2$ in frame $I_{i}$, we unproject $u_i$ to 3D point $U_i$ using the depth map $D_{i}$ and camera pose $T_{i}$, and project $U_i$ back to the camera space of frame $i+1$ though $T_{i+1}$:
\begin{equation}
    U_i = \pi^{-1}(u_i, D_{i}, T_{i}),
\end{equation}
\begin{equation}
    \hat{u}_{i+1} = \pi(U_i, T_{i+1}),
\end{equation}
here, $\pi$ is the projection operation.
The reprojection flow $\tilde{f}_{i+1} = ||\hat{u}_{i+1} - u_i||$ is the difference between the corresponding $u_i$ and $\hat{u}_{i+1}$ in pixel coordinates.
We utilize RAFT~\cite{raft} to compute the optical flow $f_{i+1}$ between $I_{i+1}$ and $I_{i}$ to supervise the reprojection flow.
Finally, we define the flow loss as:
\begin{equation}
    L_f = | | \tilde{f}_{i+1} - f_{i+1} | | ^{2}.
\end{equation}

By incorporating both the motion cues from $L_f$ and the photometric consistency from $L_c$, Rob-GS achieves robust camera pose alignment, even under challenging conditions. The final camera pose optimization process is formulated as:
\begin{equation}
\resizebox{0.6\linewidth}{!}{$
\displaystyle
    T^*_{i+1} = \arg \min_{T_{i+1}} \left\{ L_c\left( \tilde{I}_{i+1}, I_{i+1} \right) + \lambda_f L_f \right\}.
$}
\end{equation}
\ 

\subsubsection{Differentiable Camera Poses}
3DGS adopts a CUDA-based rasterizer and explicitly computes the derivatives of all Gaussian attributes to accelerate optimization. However, the rasterizer does not support the gradient propagation with respect to camera poses.
To enable efficient gradient computation for pose estimation, we derive the camera Jacobians and modify the CUDA rasterizer accordingly.

Based on Eq.~(\ref{eq:2}), the gradient of a camera pose $T_{WC}$ depends on the 2D covariance matrix $\Sigma'$ and the projected position $\mu'$ of the Gaussians.
We derive the analytical formulation from the $\alpha$-blending process.
The gradient of $T_{WC}$ with respect to the color loss $L_c$ during pose optimization is given by:
\begin{equation}
\resizebox{0.8\linewidth}{!}{$
\displaystyle
\begin{aligned}
    \frac{\partial L_c}{\partial T_{WC}} &= \frac{\partial L_c}{\partial C} \frac{\partial C}{\partial T_{WC}} \\
    &= \frac{\partial L_c}{\partial C} \left( \frac{\partial C}{\partial c_i} \frac{\partial c_i}{\partial T_{WC}} + \frac{\partial C}{\partial \alpha'_i} \frac{\partial \alpha'_i}{\partial T_{WC}} \right) \\
    &= \frac{\partial L_c}{\partial C} \left( \frac{\partial C}{\partial c_i} \frac{\partial c_i}{\partial \mu} \frac{\partial \mu}{\partial T_{WC}} + \frac{\partial C}{\partial \alpha'_i} \left(\frac{\partial \alpha'_i}{\partial \mu'}\frac{\partial \mu'}{\partial T_{WC}}+ \frac{\partial \alpha'_i}{\partial \Sigma' }\frac{\partial \Sigma'}{\partial T_{WC} }\right)\right).
\end{aligned}
$}
\end{equation}
By leveraging Lie algebra and the partial derivatives on the \( SE(3) \) manifold as described in \cite{Lie,monoGS}, we further derive the following equations:
\begin{equation}
\resizebox{0.8\linewidth}{!}{$
\displaystyle
    \frac{\partial \mu'}{\partial T_{WC}} = \frac{\partial \mu'}{\partial \mu} \frac{\mathcal{D} \mu}{\mathcal{D} T_{WC}},\ \frac{\partial \Sigma'}{\partial T_{WC}} = \frac{\partial \Sigma'}{\partial J} \frac{\partial J}{\partial \mu}\frac{\mathcal{D} \mu}{\mathcal{D} T_{WC}} + \frac{\partial \Sigma'}{\partial W} \frac{\mathcal{D} W}{\mathcal{D} T_{WC}},
$}
\end{equation}
\begin{equation}
\resizebox{0.6\linewidth}{!}{$
\displaystyle
    \begin{aligned}
    \frac{\mathcal{D}  \mu}{\mathcal{D}  T_{WC}} = \left[ I - \mu ^ { \times } \right], \ 
    \frac{\mathcal{D}  W}{\mathcal{D}  T_{WC}} = \left[ \begin{array}{ccc}
     0 & -W^ { \times }_{:,1}  \\
     0 & -W^ { \times }_{:,2}  \\
     0 & -W^ { \times }_{:,3} 
     \end{array} \right],
\end{aligned}
$}
\end{equation}
where \( \times \) denotes the skew symmetric matrix of a 3D vector, and \( W_{:,i} \) refers to the \( i \)-th column of the matrix.
Since the computation of the flow loss $L_f$ does not involve the rasterization process, we use PyTorch's Autograd for backpropagation of the gradient of $L_f$.

\subsection{Scene Segmentation \& Optimization}
\label{sec:GS optimization}

We propose an adaptive segmentation strategy that performs visibility retention check with minimal computational overhead. The check quantifies the inter-frame consistency based on the retention rate of Gaussians. When significant viewpoint changes occur, this strategy triggers segmentation updates and local optimization.

\begin{figure}[t]
  \setlength{\abovecaptionskip}{4pt}  
  \setlength{\belowcaptionskip}{0pt}
  \centering 
  \includegraphics[width=0.9\linewidth]{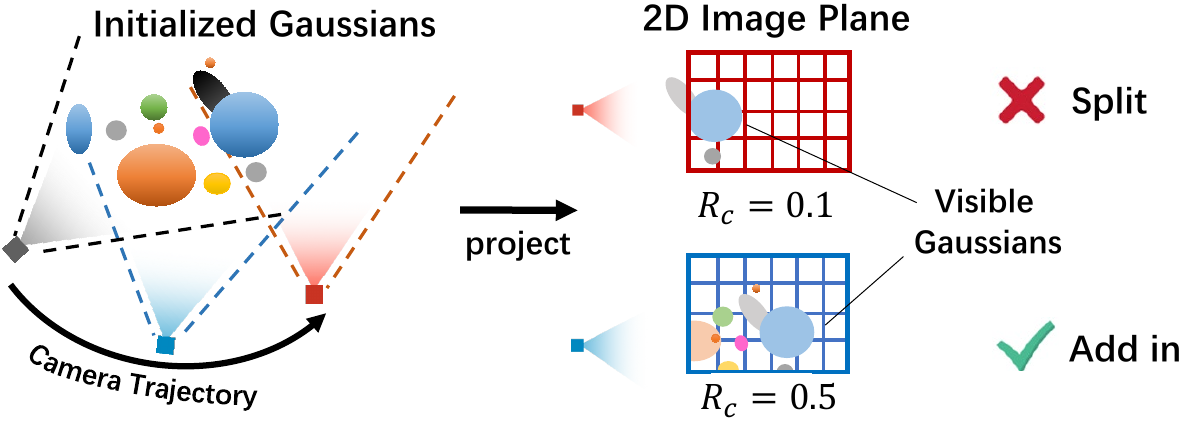}
  \caption{%
  \textbf{Gaussian visibility retention check strategy}. The blue camera exhibits a higher visibility retention rate that exceeds the threshold. Therefore, the blue camera's frame should be added to the current segment, while the red camera's frame should be assigned to a new segment.}
  \Description{ada_seg}
  \label{fig:ada_seg}
\end{figure}

\begin{table*}[t]
\centering
\setlength{\abovecaptionskip}{4pt}  
\setlength{\belowcaptionskip}{0pt}
\caption{\textbf{Novel view synthesis results on \textit{Tanks and Temples}.} The best results are highlighted in bold.}
\setlength{\tabcolsep}{6.3pt}
\footnotesize
\begin{tabular}{l|rrrr|rrrr|rrrr|rrrr}
\toprule
\multirow{2.5}{*}{\textbf{Scenes}} & \multicolumn{4}{c|}{\textbf{Nope-NeRF}} & \multicolumn{4}{c|}{\textbf{LocalRF}} & \multicolumn{4}{c|}{\textbf{CF-3DGS}} & \multicolumn{4}{c}{\textbf{Ours}} \\
\rule{0pt}{10pt}
& PSNR & SSIM & LPIPS & Time & PSNR & SSIM & LPIPS & Time & PSNR & SSIM & LPIPS & Time & PSNR & SSIM & LPIPS & Time \\
\midrule
Church & 21.80 & 0.57 & 0.57 & 144h & 27.61 & 0.83 & 0.21 & 20.7h & 21.37 & 0.64 & 0.35 & 7.7h & \textbf{30.11} & \textbf{0.90} & \textbf{0.13} & \textbf{6.5h} \\
Barn & 22.29 & 0.62 & 0.55 & 236h & 29.43 & 0.84 & 0.21 & 31.7h & 15.53 & 0.49 & 0.50 & 10.8h & \textbf{29.85} & \textbf{0.86} & \textbf{0.15} & \textbf{7.7h} \\
Museum & 22.13  & 0.60 & 0.53 & 60h & 24.76 & 0.77 & 0.27 & 8.3h & 16.61 & 0.43  & 0.46 & 4.6h & \textbf{29.43} & \textbf{0.90} & \textbf{0.12} & \textbf{2.2h} \\
Family & 21.30 & 0.58 & 0.57 & 92h & 27.14 & \textbf{0.86} & 0.18 & 11.5h & 14.54 & 0.42 & 0.50 & 6.8h & \textbf{27.15} & 0.85 & \textbf{0.15} & \textbf{2.9h} \\
Horse & 21.48 & 0.69 & 0.45 & 99h & 28.98 & 0.91  & 0.12 & 11.3h & 17.74  & 0.64  & 0.33 & 7.2h & \textbf{29.35} & \textbf{0.92} & \textbf{0.11} & \textbf{2.8h} \\
Ballroom & 20.47 & 0.51 & 0.57 & 75h & 26.25 & 0.82 & 0.20 & 10.0h & 12.71 & 0.32 & 0.50 & 6.4h & \textbf{30.23} & \textbf{0.92} & \textbf{0.09} & \textbf{2.8h} \\
Francis & 25.08 & 0.74 & 0.51 & 61h & \textbf{31.23} & 0.89 & 0.22 & 6.2h & 17.71 & 0.64 & 0.42 & 3.5h & 31.00 & \textbf{0.90} & \textbf{0.21} & \textbf{1.7h} \\
Ignatius & 18.36 & 0.39  & 0.64 & 137h & 24.79 & 0.73 & 0.27 & 16.4h & 16.36 & 0.37 & 0.46 & 8.4h & \textbf{25.25} & \textbf{0.77} & \textbf{0.19} & \textbf{5.0h} \\
\midrule
Mean & 21.61 & 0.59 & 0.55 & 113h & 27.52 & 0.83 & 0.21 & 14.5h & 16.57 & 0.49 & 0.44 & 6.9h & \textbf{29.05} & \textbf{0.88} & \textbf{0.14} & \textbf{3.9h} \\
\bottomrule
\end{tabular}
\label{tab:tnt}
\end{table*}

\begin{figure*}
    \setlength{\abovecaptionskip}{4pt}  
    \setlength{\belowcaptionskip}{0pt}
    \centering
    \includegraphics[width=0.98\linewidth]{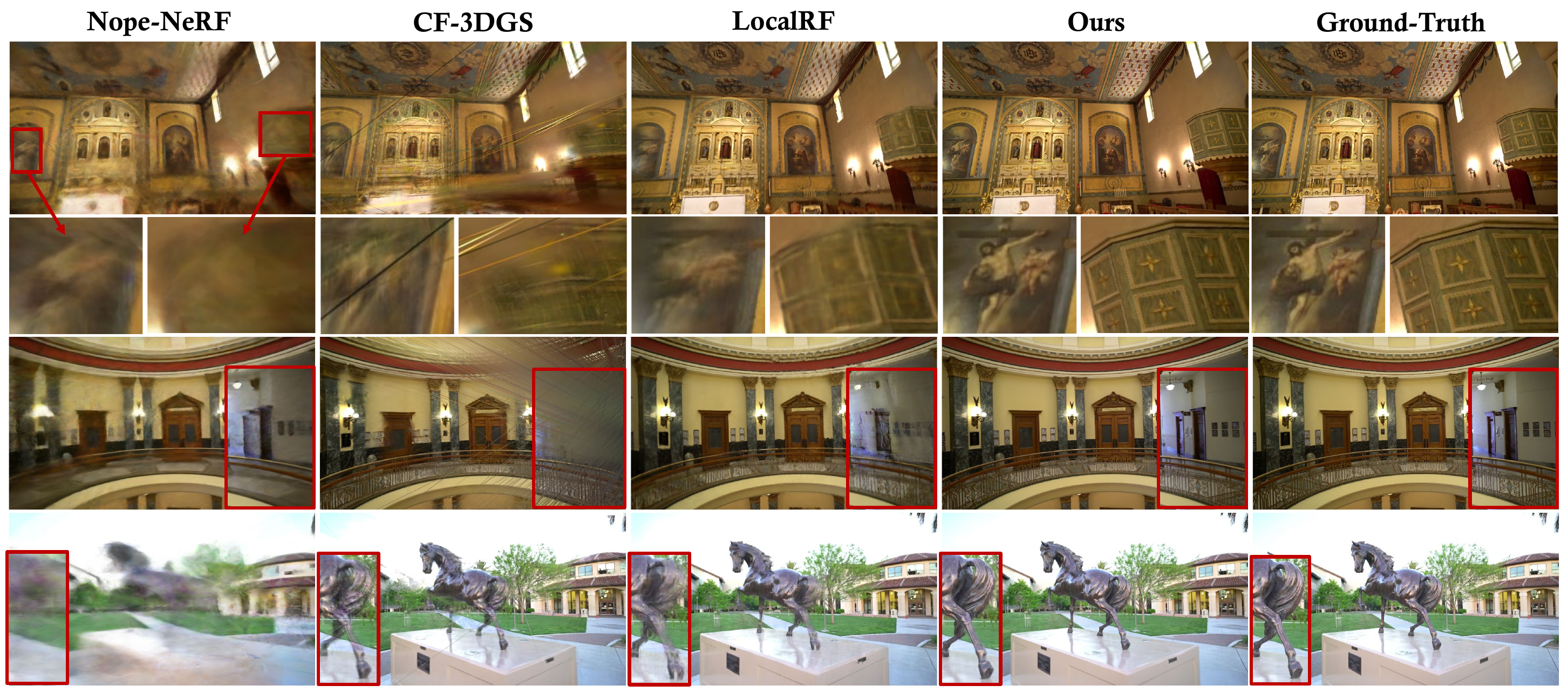}
    \caption{\textbf{Qualitative comparison for novel view synthesis on \textit{Tanks and Temples}.} Our approach produces sharper details than other baselines.}
    \Description{quali-tnt}
    \label{fig:quali-tnt}
\end{figure*}

\subsubsection{Adaptive Segmentation Strategy}
We begin with the single image fitted Gaussians $G_a$ from the last frame of the previous segment~(or the first frame at the beginning). 
We denote the Gaussian number of $G_a$ as $V_a$.
For each new frame with the estimated pose $T_i$, we project $G_a$ onto the 2D image plane using $T_i$, and count the number of visible Gaussians in this view, denoted as $V_i$ (as illustrated in Fig.~\ref{fig:ada_seg}).
A Gaussian is considered visible from a given view if it lies within the camera frustum.
Finally, the retention rate is defined as:  $R_c = V_i / V_a$.
For each newly tracked frame, we compute its retention rate $R_c$, which brings minimal computational overhead, and append it to the current segment.
The retention rate $R_c$ essentially reflects the consistency across frames. When $R_c$ falls below a predefined threshold, it indicates a significant change in scene content or viewpoint, triggering both a segmentation update and local Gaussian optimization. 
At this point, we optimize the local Gaussians using all newly tracked frames within the segment.
Under this retention check strategy, segmentation does not occur when the camera gradually zooms in, as the scene content remains consistent. On the contrary, drastic camera translation or rotation that captures a new region is more likely to initiate a segmentation update.

\subsubsection{Local Gaussian Optimization}
By adopting the above strategy, we collect a set of frames and their corresponding camera poses $\{\{I_i,T_i\}, \dots, \{I_k,T_k\}\}$ within a new segment. Subsequently, we proceed to optimize the local scene.
As illustrated in the bottom of Fig.~\ref{fig:pipeline}, we initialize the local 3DGS by lifting points from the first frame $I_i$, using its associated monocular depth map $D_i$ and camera pose $T_i$. For scenes with a large spatial extent, we additionally project points from the last frame $I_k$ to cover areas that are not captured by $I_i$.
To optimize the local 3D Gaussians, we randomly sample frames from the current segment and simultaneously minimize the color loss $L_{c}$ and depth loss $L_{d}$:
\begin{equation}
    G_{\text{local}}(\theta^{*}) = \arg \min_{\theta} \left\{  L_{c}  + \lambda_d L_{d} \right\},
\end{equation}
where $\lambda_d$ is a balancing weight. We set the number of optimization iterations based on the frame count in the current segment, and use adaptive Gaussian densification and pruning~\cite{3DGS}. Notably, we maintain a 5-frame overlap between adjacent segments during training to ensure seamless transitions.

\begin{table*}[t]
\centering
\setlength{\abovecaptionskip}{4pt}  
\setlength{\belowcaptionskip}{0pt}
\caption{\textbf{Novel view synthesis results on \textit{ScanNet} dataset.} The best results are highlighted in bold.}
\setlength{\tabcolsep}{5.7pt}
\footnotesize
\begin{tabular}{l r rrr|rrr|rrr|rrr|rrr}
\toprule
\multirow{2.5}{*}{\textbf{Methods}} & \multirow{2.5}{*}{\textbf{Times}$\downarrow$}  & \multicolumn{3}{c|}{\textbf{0231\_00}} & \multicolumn{3}{c|}{\textbf{0243\_00}} & \multicolumn{3}{c|}{\textbf{0245\_00}} & \multicolumn{3}{c|}{\textbf{0246\_00}} & \multicolumn{3}{c}{\textbf{0533\_01}} \\
\rule{0pt}{10pt}
& & PSNR & SSIM & LPIPS  & PSNR & SSIM & LPIPS & PSNR & SSIM & LPIPS & PSNR & SSIM & LPIPS & PSNR & SSIM & LPIPS \\
\midrule
Nope-NeRF & $\sim$234h & 19.13 & 0.66 & 0.64 & 27.31 & 0.85 & 0.48  & 14.49 & 0.63 & 0.64 & 24.74 & 0.83 & 0.51 & 17.00 & 0.68 & 0.65 \\
LocalRF & $\sim$36h & 29.86 & 0.89 & 0.24 & 32.89 & 0.92 & 0.19 & 30.99 & 0.91 & 0.18 & 31.67 & 0.92 & 0.20 & 30.60 & 0.88 & 0.24 \\
Ours & $\sim$\textbf{7h} & \textbf{31.63} & \textbf{0.91} & \textbf{0.20} & \textbf{33.84} & \textbf{0.93} & \textbf{0.18} & \textbf{32.46} & \textbf{0.93} & \textbf{0.17} & \textbf{32.94} & \textbf{0.93} & \textbf{0.18} & \textbf{32.45} & \textbf{0.91} & \textbf{0.21} \\
\bottomrule
\end{tabular}
\label{tab:scannet}
\end{table*}

\begin{table*}[t]
\centering
\setlength{\abovecaptionskip}{4pt}  
\setlength{\belowcaptionskip}{0pt}
\caption{\textbf{Novel view synthesis results on self-captured dataset.} The best results are highlighted in bold.}
\setlength{\tabcolsep}{5.7pt}
\footnotesize
\begin{tabular}{l r rrr|rrr|rrr|rrr|rrr}
\toprule
\multirow{2.5}{*}{\textbf{Methods}} & \multirow{2.5}{*}{\textbf{Times}$\downarrow$}  & \multicolumn{3}{c|}{\textbf{Auditorium}} & \multicolumn{3}{c|}{\textbf{Gallery}} & \multicolumn{3}{c|}{\textbf{Pavilion1}} & \multicolumn{3}{c|}{\textbf{Pavilion2}} & \multicolumn{3}{c}{\textbf{Gate}} \\
\rule{0pt}{10pt}
& & PSNR & SSIM & LPIPS  & PSNR & SSIM & LPIPS & PSNR & SSIM & LPIPS & PSNR & SSIM & LPIPS & PSNR & SSIM & LPIPS \\
\midrule
Nope-NeRF & $\sim$107h & 21.01 & 0.51 & 0.58 & 19.89 & 0.66 & 0.52  & 17.03 & 0.38 & 0.62 & 20.95 & 0.58 & 0.52 & 23.77 & 0.62 & 0.53 \\
LocalRF & $\sim$15h & 25.18 & 0.74 & 0.27 & 22.57 & 0.77 & 0.27 & 20.88 & 0.67 & 0.33 & 23.91 & 0.79 & 0.24 & 28.04 & 0.84 & 0.19 \\
Ours & $\sim$\textbf{4h} & \textbf{28.28} & \textbf{0.85} & \textbf{0.13} & \textbf{25.42} & \textbf{0.84} & \textbf{0.17} & \textbf{23.36} & \textbf{0.80} & \textbf{0.17} & \textbf{28.83} & \textbf{0.92} & \textbf{0.08} & \textbf{29.70} & \textbf{0.89} & \textbf{0.12} \\
\bottomrule
\end{tabular}
\label{tab:self_capture}
\end{table*}

\begin{figure*}
    \setlength{\abovecaptionskip}{4pt}  
    \setlength{\belowcaptionskip}{0pt}
    \centering
    \includegraphics[width=0.96\linewidth]{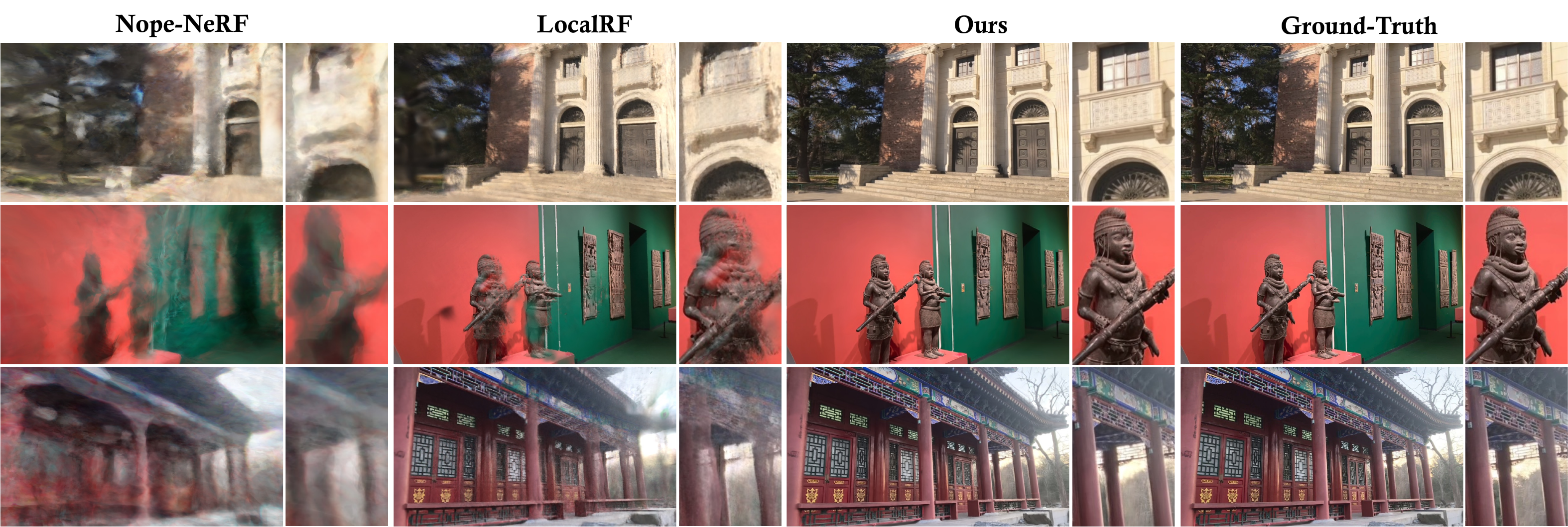}
    \caption{\textbf{Qualitative comparison for novel view synthesis on self-captured dataset.} Our approach produces the most realistic rendering results among other baselines. Highly recommend zooming in for better comparison.}
    \Description{quali-capture}
    \label{fig:quali-capture}
\end{figure*}

\section{Experiments}
\subsection{Experimental Setup}
\subsubsection{Datasets \& Baselines}
\dzh{We conduct comprehensive experiments on three datasets, including \textit{Tanks and Temples}~\cite{tanks}, \textit{ScanNet}~\cite{scannet}, and our self-captured dataset.}
Below, we briefly summarize each dataset. For more details, please refer to our supplementary materials.
\textbf{\textit{Tanks and Temples.}} This dataset contains video sequences captured using stabilizers, featuring diverse environments such as large indoor scenes, outdoor sculptures and buildings. We evaluate pose accuracy and novel view synthesis quality on 8 scenes from this dataset. Unlike previous works~\cite{nope-nerf,CF3dgs} that focus on small, limited regions, we use much longer sequences that cover much larger areas.
\textbf{\textit{ScanNet.}} We select 5 scenes to evaluate novel view synthesis quality in room-scale indoor environments. For each scene, we test with different frame sampling rates, resulting in 1000-2000 input images.
\textbf{\textit{Self-Captured Dataset.}} The dataset consists of 5 scenes casually captured using a handheld mobile phone in an unconstrained manner, without any stabilizing equipment. Unlike curated datasets, these self-captured videos reflect realistic capturing conditions commonly encountered in daily life. The scenes feature roaming trajectories in a museum gallery (Gallery), circular movements around a building (Pavilion), and random wandering paths (Auditorium, Gate). Compared to \textit{Tanks and Temples} and \textit{ScanNet}, this dataset has both more challenging camera trajectories, more complex scene layouts, and richer texture variations, making it a valuable benchmark for evaluating robustness and generalization capabilities of different methods.
We visualize the camera trajectories of some scenes from this newly collected dataset in Fig.~\ref{fig:teaser} and Fig.~\ref{fig:supp-traj}.

We compare Rob-GS with a representative pose-free NeRF method (Nope-NeRF~\cite{nope-nerf}), a method designed for reconstructing radiance fields from long video sequences (LocalRF~\cite{localrf}), and the first SfM-free 3DGS method (CF-3DGS~\cite{CF3dgs}). For all datasets, we reserve one out of every eight video frames for novel view synthesis evaluation.

\subsubsection{Evaluation Metrics}
We use PSNR, SSIM, and LPIPS to evaluate the quality of novel view synthesis.
Since we do not have the ground-truth poses of the test views, following CF-3DGS~\cite{CF3dgs}, we freeze the trained 3DGS and directly optimize the camera poses by minimizing the differences between rendered images and test views, and test under the optimized poses.

We compare our estimated poses with the COLMAP poses. Since COLMAP are not accurate in the long video setting, we report the Relative Pose Error (RPE)~\cite{rpe} metric.
In addition, we report the training time of all methods to show the efficiency.

\subsubsection{Implementation Details}
We implement our Rob-GS using PyTorch framework. All experiments are conducted on a single NVIDIA A6000 GPU.
During adjacent pose estimation, we first fit the single image Gaussians for 500 iterations, which typically takes 7 seconds. Then, we track the camera pose for 250 iterations with an early-stop mechanism, which terminates if the pose update falls below $10^{-4}$. This tracking step generally completes within 1 second.
For local segment optimization, the total number of iterations is set to 300 times the segment length. Starting from 500 iterations, we densify the Gaussians every 100 iterations and stop densification once reaching 70\% of the total optimization iterations.
More details can be found in the supplementary materials.

\subsection{Quantitative Comparison}
We present the quantitative results in Tables~\ref{tab:tnt}–\ref{tab:self_capture} for the \textit{Tanks and Temples}, \textit{ScanNet}, and our self-captured datasets.
For \textit{Tanks and Temples} dataset, both Nope-NeRF and CF-3DGS exhibit inferior rendering quality 
when handling long sequences. Additionally, Nope-NeRF suffers from lengthy training time.
On the \textit{ScanNet} and self-captured datasets, CF-3DGS frequently fails due to out-of-memory (OOM) issue, primarily caused by severely inaccurate pose estimations. These erroneous poses lead to substantially inflated photometric losses, which in turn trigger excessive densification of 3D Gaussians during scene optimization, eventually resulting in memory overflow. Hence, we do not report CF-3DGS results on these two datasets.

Our Rob-GS achieves superior rendering quality and faster training speed across all datasets. Compared to LocalRF, our approach shows significant improvements while using only 25\% of the training time. On the \textit{Tanks and Temples} dataset, we improve the average PSNR by 1.53 dB and reduce LPIPS by 30\%. On the \textit{ScanNet} dataset, our method improves the average PSNR by 1.46 dB. For the more challenging self-captured dataset, Rob-GS surpasses LocalRF in terms of PSNR, SSIM, and LPIPS by a large margin.

As ground-truth poses are not available, we use COLMAP poses as the pseudo ground-truth, and report RPE on \textit{Tanks and Temples} dataset as a reference for pose estimation quality.

\begin{table}
\centering
\setlength{\abovecaptionskip}{0pt}  
\setlength{\belowcaptionskip}{4pt}
\setlength{\tabcolsep}{16pt}
\begin{tabular}{lrrr}
\toprule
Metric  & LocalRF & CF-3DGS  & Ours \\
\midrule
$\text{RPE}_{t}\downarrow$ & 0.110 & 0.019 & \textbf{0.015}  \\
$\text{RPE}_{r}\downarrow$ & 0.369  & 0.161 & \textbf{0.149} \\
\bottomrule
\end{tabular}
\caption{\textbf{Pose comparisons on \textit{Tanks and Temples}.} We directly use the COLMAP poses as the pseudo ground-truth. The unit of $\text{RPE}_{r}$ is in degrees, and $\text{RPE}_{t}$ is scaled by 100.}
\label{tab:pose_eval}
\end{table}

\begin{figure}[t]
  \setlength{\abovecaptionskip}{4pt}  
  \setlength{\belowcaptionskip}{0pt}
  \centering 
  \includegraphics[width=\linewidth]{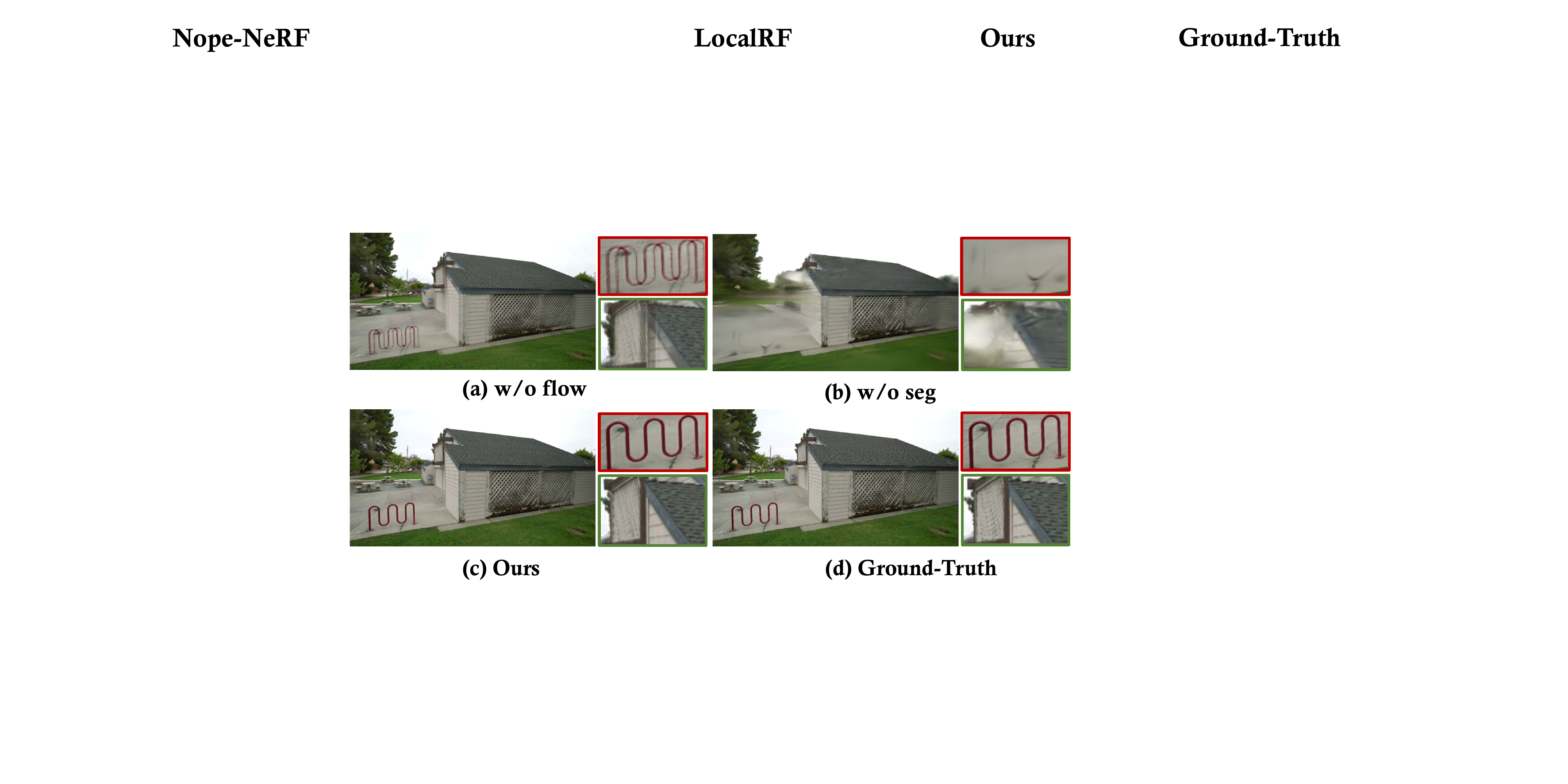}
  \caption{%
  \textbf{The visualizations of our ablation study.} Artifacts will appear when any of our proposed components are removed.}
  \Description{ablation}
  \label{fig:ablation}
\end{figure}

\subsection{Qualitative Comparison}
Fig.~\ref{fig:quali-tnt} compares rendering results on the \textit{Tanks and Temples} dataset, and Fig.~\ref{fig:quali-capture} shows results on the self-captured dataset.
Our Rob-GS can produce high-fidelity and sharp renderings throughout the entire trajectory (please refer to our demo video).
In contrast, CF-3DGS generates large amounts of needle-like artifacts, which severely occlude the scene.
Nope-NeRF generates overly blurry results in the entire image, and sometimes produces renderings with misaligned poses (as shown in the first row, first column of Fig.~\ref{fig:quali-capture}).
LocalRF struggles to render fine details, such as distant objects (\textit{e.g.,} trees appearing as blurry mess) and intricate textures.
Thanks to the robust pose estimation and local Gaussian optimization within each segment, our method consistently produces sharp results among all test views, even in very challenging settings.

\subsection{Ablation Study}
We conduct ablation study to validate our design choices. Specifically, we examine the impact of the flow loss in the adjacent pose tracking module (w/o flow) as well as our local segment optimization scheme (w/o seg). Results are shown in Tab.~\ref{tab:ablation} and Fig.~\ref{fig:ablation}.

\subsubsection{Effectiveness of Flow Loss}
In our experiments, we observe that inaccurate camera poses lead to ghosting artifacts. 
This arises because incorrect poses break the geometric consistency required for accurate ray intersections, causing rays to fail to converge at the correct 3D locations.
Consequently, the model learns an incorrect 3D structure and produces visual artifacts such as duplicated edges and blurred textures, as illustrated in Fig.~\ref{fig:ablation}(a). Although this issue may have a limited impact on PSNR metrics, the improvements brought by the flow loss are clearly reflected in visual quality, particularly in challenging scenes with large and complex camera motion (demonstrated in Fig.~\ref{fig:ablation}(c)).

\subsubsection{Local Segment Optimization vs Global Optimization}
We validate the effectiveness of our segment-wise optimization scheme by replacing it with a global 3DGS training strategy. Specifically, we first estimate the poses for all training views and then perform global optimization of the 3D Gaussians, following the procedure of Kerbl \textit{et al.}~\shortcite{3DGS}. Fig.~\ref{fig:ablation}(b) shows that the global optimization scheme struggles to render fine-grained details in the scene. In contrast, our local segment optimization scheme is crucial for producing high-quality results. Furthermore, it is also better suited for arbitrarily long input sequences, as it enables incremental updates and seamless integration with previously reconstructed content. Compared to global optimization, it avoids the tedious pre-computing of all camera poses, and local results remain unaffected by subsequent inputs. Hence, local optimization mitigates the catastrophic forgetting problem commonly encountered in large-scale scenarios, enhancing the robustness and scalability of our approach.

\begin{table}
\centering
\setlength{\abovecaptionskip}{0pt}  
\setlength{\belowcaptionskip}{4pt}
\setlength{\tabcolsep}{14pt}
\begin{tabular}{lrrr}
\toprule
Configs  & PSNR$\uparrow$ & SSIM$\uparrow$ & LPIPS$\downarrow$ \\
\midrule
Ours w/o flow & 28.53 & 0.86 & 0.15 \\
Ours w/o seg & 21.12 & 0.65 & 0.34 \\
Ours full & \textbf{29.05}  & \textbf{0.88}  & \textbf{0.14} \\
\bottomrule
\end{tabular}
\caption{\textbf{Ablation study results on \textit{Tanks and Temples} dataset.} The best results are highlighted in bold.}
\label{tab:ablation}
\end{table}

\section{Conclusion \& Limitations}
We present Rob-GS, a robust framework for progressively estimating camera poses and optimizing 3D Gaussians from arbitrarily long video sequences. Our method introduces a novel flow-induced adjacent pose tracking module for stable pose estimation under challenging, casually captured camera trajectories, and an adaptive segmentation strategy based on Gaussian visibility retention for efficient local optimization. We conduct extensive experiments on three real-world datasets. The results show that Rob-GS significantly enhances the robustness and rendering quality over all baselines, demonstrating scalability and practical applicability.

Currently, our method does not incorporate loop closure, as it is often impractical for arbitrarily long, casually captured videos in which loops may be absent or difficult to detect. Nonetheless, our segment-wise optimization maintains stable rendering quality even for long and unconstrained sequences. We regard how to effectively detect and accordingly incorporate loop closure strategy into our method as a future research direction.

\begin{figure*}[htbp]
    \centering
    \setlength{\abovecaptionskip}{4pt}  
    \setlength{\belowcaptionskip}{0pt}
    \includegraphics[width=0.96\linewidth]{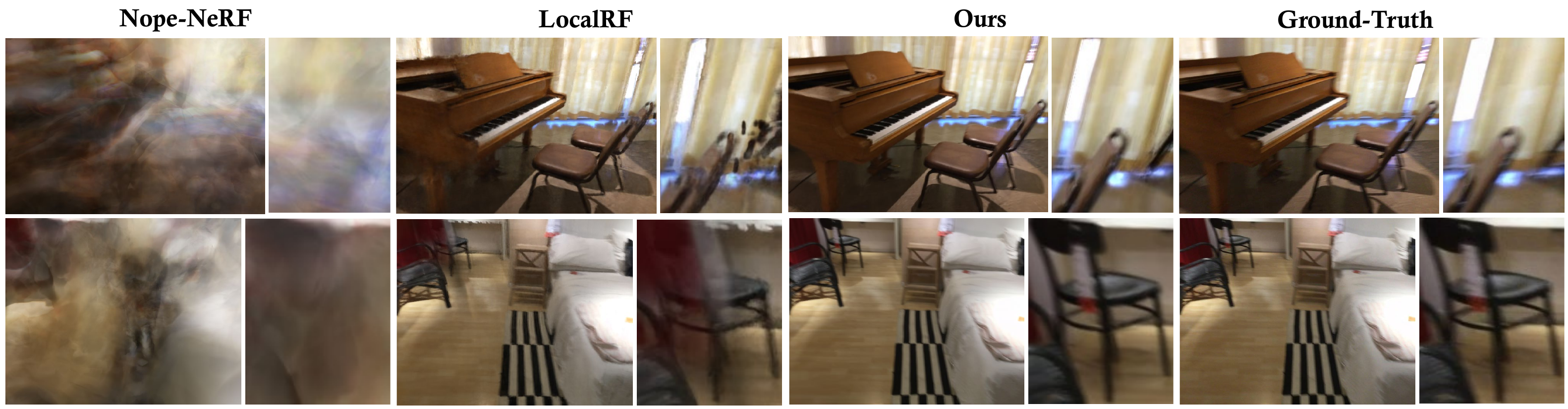}
    \caption{\textbf{Qualitative results of novel view synthesis on \textit{ScanNet} dataset}. Our Rob-GS is able to preserve fine-grained datails.}
    \Description{scan-res}
    \label{fig:scan-res}
    \vspace{0.15cm}
\end{figure*}

\begin{figure*}[htbp]
    \centering
    \setlength{\abovecaptionskip}{4pt}  
    \setlength{\belowcaptionskip}{0pt}
    \includegraphics[width=0.96\linewidth]{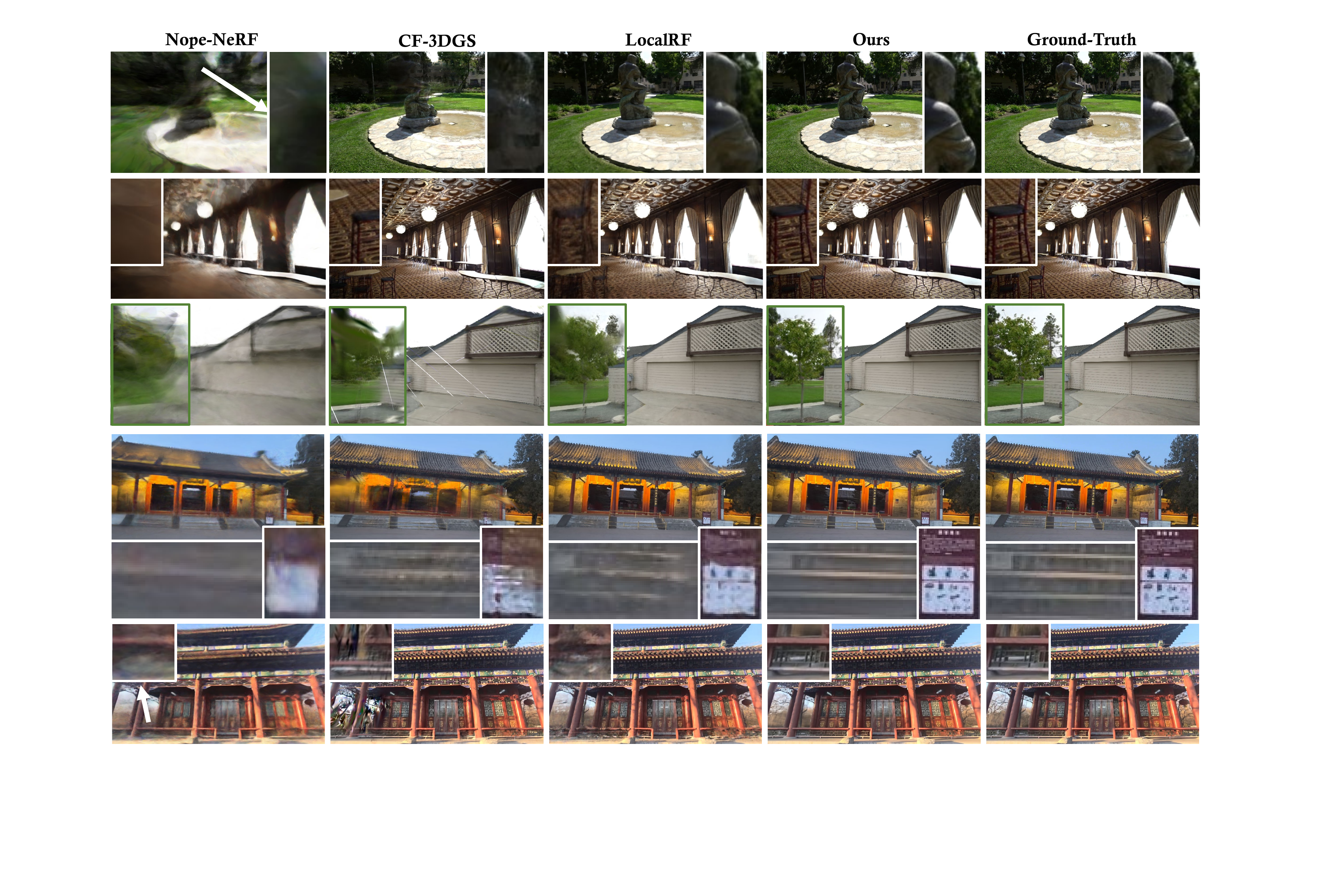}
    \caption{\textbf{Qualitative results of novel view synthesis on \textit{Tanks and Temples} and self-captured datasets}. Zoom in for better comparison.}
    \Description{supp-quality}
    \label{fig:supp-quality}
    \vspace{0.25cm}
\end{figure*}

\begin{figure*}[htbp]
    \centering
    \setlength{\abovecaptionskip}{4pt}  
    \setlength{\belowcaptionskip}{0pt}
    \includegraphics[width=0.97\linewidth]{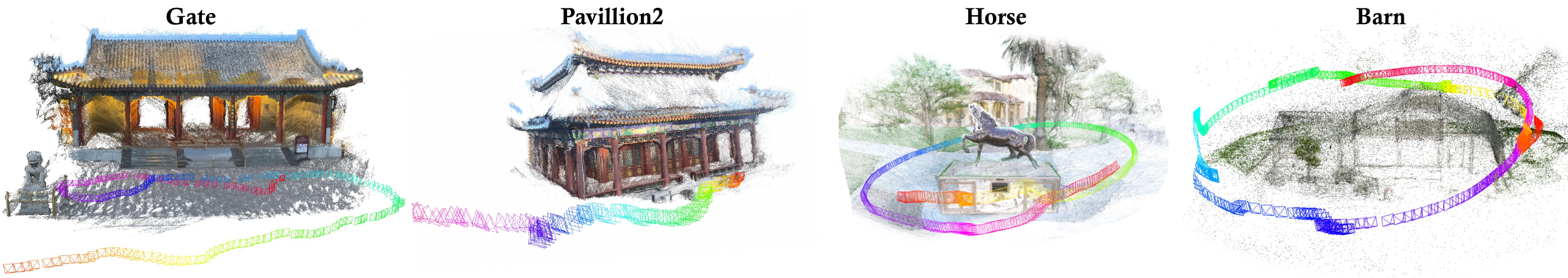}
    \caption{\textbf{Camera trajectories estimated by Rob-GS}. The first two scenes are from self-captured dataset, the latter two from \textit{Tanks and Temples}.}
    \Description{supp-traj}
    \label{fig:supp-traj}
\end{figure*}

\balance

\normalem

%
%
%

\bibliographystyle{ACM-Reference-Format}
\bibliography{sample-bibliography}

\end{document}